%% file: main.tex
\documentclass{article} % For LaTeX2e
\usepackage[T1]{fontenc}
\usepackage{iclr2027_conference,times}

\input{math_commands.tex}

\usepackage{hyperref}
\usepackage{url}
\usepackage{graphicx}
\usepackage{float}
\usepackage{booktabs}
\usepackage{array}
\usepackage{xcolor}
\newcommand{\cbox}[2][1.5ex]{\textcolor{#2}{\rule{#1}{#1}}}
\definecolor{tokenblue}{RGB}{68,114,196}

\usepackage{multirow}
\usepackage{multicol}
\usepackage{colortbl}
\usepackage{adjustbox}
\usepackage{amsmath}
\usepackage{amssymb}
\usepackage{mathtools}
\usepackage{amsthm}

\usepackage{pifont}
\usepackage{caption}
\usepackage{makecell}
\usepackage{tabularx}
\usepackage{rotating}
\usepackage{setspace}

\title{S2D-Splat: Static-First Gaussian Reconstruction of Dynamic Driving Scenes with Layer-Wise Feature Guidance}
\author{
\parbox[t]{0.9\textwidth}{\centering
Jinheng Li\textsuperscript{1*},
Tianchen Deng\textsuperscript{2*$\dagger$},
Bufan Zhang\textsuperscript{3},
Qu Chen\textsuperscript{4$\dagger$},
Jianwu Fang\textsuperscript{1$\dagger$}
}\\
\\
\parbox[t]{0.9\textwidth}{\centering
Xi'an Jiaotong University \textsuperscript{1}, Shanghai Jiao Tong University \textsuperscript{2},\\ ShanghaiTech University\textsuperscript{3},MEGVII Technology{\textsuperscript{4}
}\\
\textsuperscript{*}Equal contribution. \textsuperscript{$\dagger$}Corresponding authors: dengtianchen@sjtu.edu.cn, bob360068284@gmail.com, fangjianwu@mail.xjtu.edu.cn.}
}

\iclrfinalcopy % arXiv: show real authors, no line numbers
\begin{document}

\maketitle

\input{tex/0_Abstract}

\begin{figure*}[h]
    \centering
    \includegraphics[width=\textwidth]{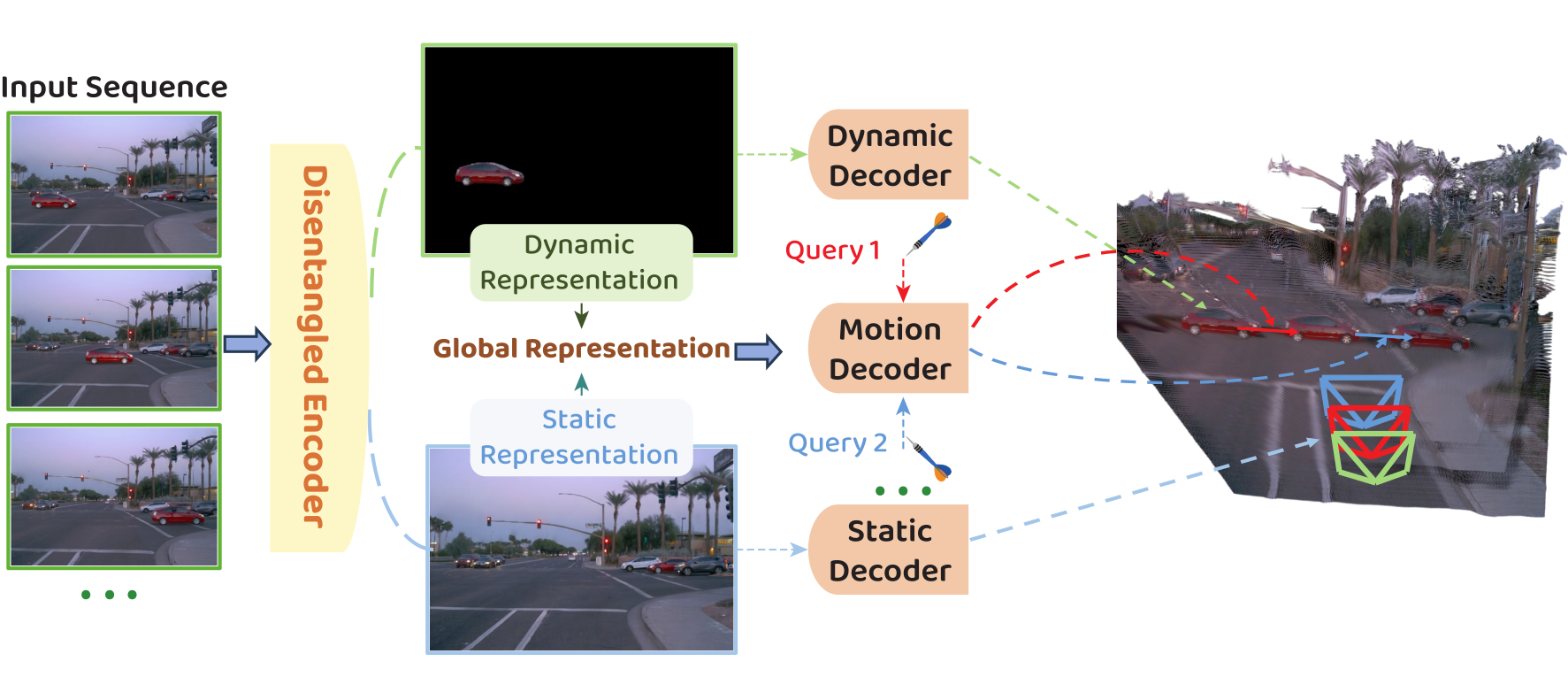}
    \caption{We propose S2D-Splat, a feedforward Gaussian reconstruction framework that achieves precise dynamic-static decoupling and instance-level motion prediction for complex driving scenes. It disentangles multi-frame input into static and dynamic representations, reconstructs each representation independently, and fuses them into a global representation that guides a motion decoder to recover accurate and temporally consistent per-instance 3D trajectories (e.g., Query~1, Query~2).}
    \label{fig:intro}
\end{figure*}

\input{tex/1_Introduction}
\input{tex/2_Related_Work}
\input{tex/3_Method}

\input{tex/4_Experiments}

\input{tex/5_Conclusion}

\clearpage
\newpage

\subsection*{AI Use Statement}
In this work, we used generative AI tools, including GPT-5.4 and Claude Sonnet 5, for writing assistance, text polishing, code debugging, and support during the evaluation process. All reported metrics and derived comparisons were computed directly from experimental evaluations, not estimated or fabricated by generative AI. We have reviewed all AI-assisted content and take responsibility for the final text, claims, and artifacts presented in this paper.

\bibliography{iclr2027_conference}
\bibliographystyle{iclr2027_conference}

\input{tex/6_Appendix}

% \appendix
% \section{Appendix}
% You may include other additional sections here.

\end{document}

%% file: math_commands.tex
\usepackage{amsmath,amsfonts,bm}

\def\eqref#1{equation~\ref{#1}}
\def\1{\bm{1}}

\DeclareMathAlphabet{\mathsfit}{\encodingdefault}{\sfdefault}{m}{sl}
\SetMathAlphabet{\mathsfit}{bold}{\encodingdefault}{\sfdefault}{bx}{n}

%% file: tex/0_Abstract.tex
\vspace{-10pt}
\begin{abstract}
\label{sec:abstract}
\vspace{-10pt}

Moving objects introduce cross-frame inconsistencies that complicate camera estimation and static geometry recovery in driving scenes. Recent feed-forward Gaussian methods explicitly separate static and dynamic primitives, but often estimate camera parameters and geometry from shared features that jointly encode both components, leaving static reconstruction susceptible to dynamic interference. To address this challenge, we propose \textbf{S2D-Splat}, a feed-forward framework for \textbf{static-first Gaussian reconstruction} with \textbf{layer-wise feature guidance}. Our framework separates static and dynamic features before camera and geometry estimation, enabling the static branch to recover camera parameters and static 3D Gaussians with reduced dynamic interference. Intermediate static features are cached at each layer and reused as fixed conditioning at corresponding layers of the dynamic branch. This asymmetric information flow preserves the established static representations while providing stable geometric context for reconstructing dynamic 3D Gaussians. An instance-level motion decoder further queries the fused static and dynamic representations to estimate object trajectories, supporting motion-aligned Gaussian fusion and temporal interpolation. Experiments on Waymo and zero-shot evaluations on nuScenes and the MM-AU accident dataset demonstrate that S2D-Splat outperforms existing approaches on most evaluated metrics and generalizes to highly dynamic pre-collision driving scenes.

\end{abstract}

%% file: tex/1_Introduction.tex
\section{Introduction}
\label{sec:introduction}

Reconstructing dynamic driving scenes from image sequences provides a geometric basis for autonomous driving simulation, accident analysis, and near-miss replay. A useful reconstruction must recover camera motion, preserve the static environment, and represent moving agents coherently in a common 3D space. These requirements become particularly challenging in dense traffic and accident scenarios, where abrupt object motion and changing occlusions disrupt cross-frame correspondences. Errors in camera estimation can then distort static geometry and confound object motion, while fragmented dynamic reconstructions compromise the fidelity of scene replay. Reliable reconstruction therefore requires carefully managing the interaction between static scene recovery and dynamic object modeling.

Per-scene optimization has enabled high-quality reconstruction through methods such as 3DGS~\citep{kerbl3Dgaussians}, EmerNeRF~\citep{yang2023emernerf}, and DeformableGS~\citep{yang2024deformable}. However, substantial scene-specific computation and reliance on camera poses limit their scalability and complicate their application to uncalibrated driving videos. Feed-forward Gaussian methods address the computational burden by directly predicting renderable 3D representations from input images~\citep{pixelsplat,mvsplat}. Approaches such as MVSplat~\citep{mvsplat}, NoPoSplat~\citep{noposplat}, and DepthSplat~\citep{depthsplat} primarily target static scenes, leaving moving objects challenging to reconstruct consistently. STORM~\citep{yang2024storm} extends feed-forward reconstruction to dynamic driving scenes through motion-based Gaussian aggregation, but requires known camera parameters and operates within short temporal windows.

More recent methods, including DGGT~\citep{dggt} and StreetForward~\citep{streetforward}, support reconstruction from unposed images and explicitly distinguish static and dynamic Gaussian primitives. DGGT predicts dynamic masks to partition Gaussian maps, while StreetForward uses motion-aware features to predict velocities and dynamic probabilities for Gaussian separation and temporal aggregation. Despite their dedicated dynamics modules, camera parameters and Gaussian geometry are estimated from backbone features that jointly aggregate static and dynamic image content. Their Gaussian-level decomposition therefore does not explicitly isolate the static features used by these earlier estimation stages. Moving objects can still influence the representations from which camera parameters and static geometry are recovered, even when the resulting Gaussian primitives are subsequently assigned to separate components.

This observation points to an asymmetric dependency between static and dynamic reconstruction. Static scene recovery benefits from suppressing interference from moving objects, whereas dynamic reconstruction benefits from the geometric reference provided by the static environment. Early feature separation addresses the first requirement, but effective reconstruction also requires controlled interaction between the separated representations. In particular, static information should remain available to the dynamic branch without allowing dynamic updates to modify the established static representations. This motivates a static-first design that combines feature separation before camera and geometry estimation with directed, layer-wise guidance from static to dynamic features.

To this end, we propose \textbf{S2D-Splat}, a feed-forward framework for \textbf{static-first Gaussian reconstruction} of dynamic driving scenes with \textbf{layer-wise feature guidance}. As illustrated in Figure~\ref{fig:intro}, an initial encoder extracts image features, which are partitioned using instance masks obtained with SAM2~\citep{ravi2025sam}. A feature separator uses masked attention to restrict interactions from dynamic regions and deformable attention to refine dynamic features and recover static cues from mixed boundary tokens. By separating features before camera estimation and Gaussian reconstruction, the framework reduces dynamic interference during the recovery of static scene structure.

The static branch processes the separated features through an alternating-attention backbone~\citep{vggtomega} to estimate camera parameters and reconstruct static 3D Gaussians. Its intermediate static features are cached at each layer and subsequently reused to guide the dynamic branch. Specifically, at every layer of dynamic reconstruction, static token positions are replaced with the corresponding cached static tokens, while dynamic tokens are updated through attention to the combined representation. This mechanism preserves the static representations established at each layer while allowing dynamic features to draw on their geometric context. The recovered camera parameters further place static and dynamic Gaussians in a common coordinate system, connecting the two reconstruction branches through a consistent geometric reference.

To recover object motion, we introduce an instance-level decoder that queries the fused static and dynamic representations. Each query specifies an instance and timestep, enabling selective decoding of instance poses. The estimated poses provide shared transformations for aligning an instance's Gaussians across frames, allowing complementary observations to be fused in a common object reference frame. This supports temporal interpolation and coherent reconstruction of an object's overall motion under partial observations, extending the static-first geometry framework into a dynamic scene representation.

We train S2D-Splat on the Waymo Open Dataset~\citep{sun2020scalability} and evaluate its zero-shot generalization on nuScenes~\citep{caesar2020nuscenes} and the MM-AU accident dataset~\citep{fang2024mmau}. Experiments demonstrate improved rendering quality and depth accuracy on Waymo, together with strong reconstruction and temporal interpolation performance on nuScenes. Evaluation on MM-AU further examines generalization to pre-collision clips of accident scenarios beyond the training distribution. Ablation studies investigate the effects of static-first decoupling, asymmetric layer-wise feature guidance, and instance-level motion decoding.

Our main contributions are:
\begin{itemize}
\item We introduce a static-first feed-forward Gaussian reconstruction framework that separates static and dynamic features before camera and geometry estimation, reducing dynamic interference during static scene recovery.
\item We propose asymmetric layer-wise feature guidance that reuses cached static features as fixed conditioning for the dynamic branch, preserving the established static representations while providing geometric context for dynamic reconstruction.
\item We develop an instance-level motion decoder that queries the fused scene representation to estimate object poses, enabling motion-aligned Gaussian fusion across frames and temporal interpolation.
\item We demonstrate the effectiveness of S2D-Splat on Waymo and its zero-shot generalization to nuScenes and MM-AU, covering both conventional driving scenes and pre-collision accident clips.
\end{itemize}

%% file: tex/2_Related_Work.tex
\vspace{-5pt}
\section{Related Work}
\label{sec:related_work}
\vspace{-5pt}

\paragraph{Feed-forward Geometry and Gaussian Reconstruction.}
Neural 3D scene representations support diverse robotic tasks~\citep{deng2025best3dscenerepresentation}; Gaussian representations further enable large-scale SLAM~\citep{deng2025vpgs} and driving world models~\citep{Deng_2026_CVPR,deng2026gaussiandwm++}. NeRF-based methods~\citep{deng2026prosgnerf,yang2023emernerf} and Gaussian-based driving reconstruction approaches~\citep{zhou2024drivinggaussian,yan2024streetgaussians} achieve high visual fidelity through per-scene optimization, but incur substantial per-scene computation. Feed-forward methods instead learn transferable reconstruction priors across scenes. DUSt3R and VGGT establish generalizable geometric representations~\citep{dust3r,vggt}, while VGGT-$\Omega$ improves reconstruction across static and dynamic scenes through scaled training~\citep{vggtomega}. Driving-specific models further incorporate sensor configurations and scene priors~\citep{drivevggt,dvgt,dvgt2}. In parallel, generalizable Gaussian methods directly predict renderable primitives from one or a few images~\citep{depthsplat,noposplat,flare,anysplat,gslrm,flash3d}. For driving scenes, limited camera overlap and incomplete visibility motivate joint depth--pose learning~\citep{drivingforward}, complementary pixel--volume representations~\citep{omniscene}, and volumetric fusion~\citep{evolsplat}. Sparse LiDAR observations~\citep{adgaussian} and pretrained geometry models~\citep{vgd,vggd} provide additional geometric constraints. 
%These advances establish the geometric and rendering priors our static-first reconstruction framework builds on.

\paragraph{Static--Dynamic Separation and Feature Guidance.}
Existing methods address static--dynamic interaction at different stages of reconstruction. DrivingRecon~\citep{drivingrecon} separates static and dynamic rendering supervision to exploit temporal observations. DGGT~\citep{dggt} predicts dynamic masks to partition Gaussian maps, while StreetForward~\citep{streetforward} constructs motion-aware features through causal attention and predicts velocities and dynamic probabilities for Gaussian separation and temporal aggregation. In both methods, camera parameters and Gaussian geometry are recovered from backbone features that jointly aggregate static and dynamic image content without explicitly static features isolation. Other approaches intervene earlier. MUT3R~\citep{shen2025mut3r} derives motion cues from attention maps and suppresses dynamic interference in early transformer layers without retraining. DrivingScene~\citep{drivingscene} adopts a staged learning strategy that first establishes a static scene prior and subsequently trains a residual flow network with the static backbone frozen. S2D-Splat combines feature separation before camera and geometry estimation with directed, layer-wise static-to-dynamic guidance. After recovering camera parameters and static Gaussians, it reuses cached intermediate static features as fixed conditioning at corresponding layers of the dynamic branch. This preserves the established static representations while allowing dynamic tokens to use their geometric context throughout reconstruction.

\paragraph{Motion Modeling in Dynamic Scenes.}
Feed-forward dynamic reconstruction methods represent temporal changes through several complementary mechanisms. STORM, Flux4D, and UFO-4D predict motion fields to align dynamic geometry~\citep{yang2024storm,flux4d,ufo4d}. ReconDrive models Gaussian motion via segment-wise velocities~\citep{recondrive}, while Instant NuRec predicts piecewise-linear trajectories~\citep{instantnurec}. UniSplat fuses observations within a 3D latent scaffold~\citep{unisplat}, and Ground4D performs localized temporal Gaussian aggregation for off-road scenes~\citep{ground4d_offroad}. Object-centric approaches offer another form of temporal organization: UFO uses object poses to guide dynamic modeling~\citep{ufo}, while PointForward uses 3D boxes and scene graphs for instance-level motion propagation~\citep{pointforward}. D4RT~\citep{d4rt} introduces a query-based formulation that independently decodes point positions across time from a global spatio-temporal representation, supporting both sparse and dense queries. Following this formulation, our motion decoder operates at the instance level, by querying fused static and dynamic features to estimate a selected instance's pose at a requested timestep. The resulting transformations align its Gaussians across observations, supporting motion-aligned fusion and temporal interpolation.

%% file: tex/3_Method.tex
\section{S2D-Splat}
\label{sec:method}

\begin{figure*}[t]
    \centering
    \includegraphics[width=\textwidth]{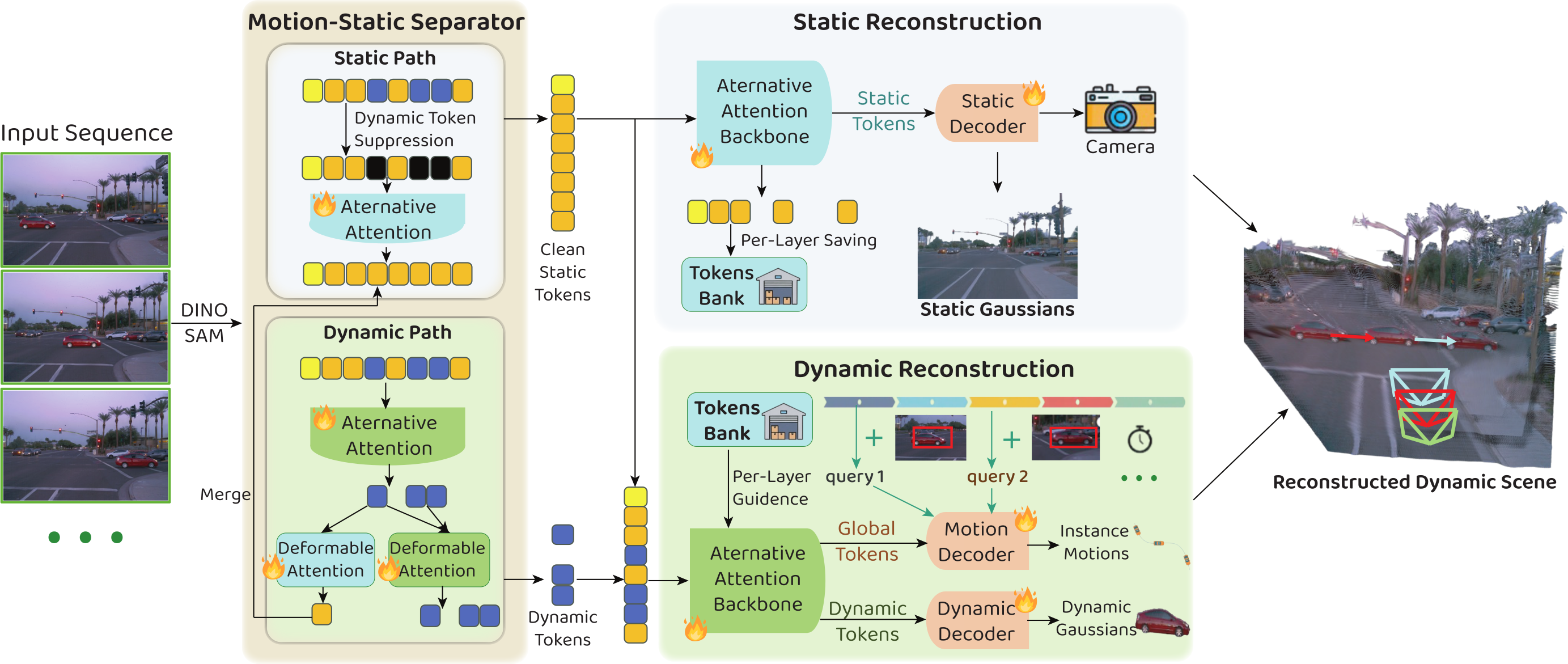}
    \caption{Overview of \textbf{S2D-Splat}. Given unposed images, we encode each frame into patchified features and use SAM2 to classify tokens as static or dynamic, which a Motion-Static Separator then refines into clean static and dynamic tokens (camera in \cbox{yellow!80}, static in \cbox{orange!60}, dynamic in \cbox{tokenblue}, suppressed dynamic in \cbox{black}). The static tokens pass through alternating attention and are decoded by heads into camera parameters and a static Gaussian field, with each layer's camera and static tokens cached in a Tokens Bank. The dynamic reconstruction mirrors this but takes the fused static and dynamic tokens as input, re-fusing the extracted dynamic tokens with the Tokens Bank after each layer to yield global tokens and a dynamic Gaussian field. From the global tokens, instance-specific queries retrieve each instance's motion, and all components merge to form the full scene reconstruction.}
    \label{fig:pipeline}
\end{figure*}
We propose \textbf{S2D-Splat}, a pose-free feedforward reconstruction framework that achieves decoupling of static and dynamic reconstruction. To prevent dynamic noise from contaminating the static reconstruction, we first design a Motion-Static Separator that decouples the dynamic and static features (\S~\ref{sec:separator}). Building on these separated features, we adopt a two-stage reconstruction strategy that constructs the static and dynamic Gaussian fields separately: once static reconstruction is shielded from dynamic objects, the dynamic objects are in turn reconstructed with reference to this clean static scene information (\S~\ref{sec:recon}). The framework is \textbf{static-first}: separation precedes camera and geometry estimation, and the two stages are coupled by \textbf{asymmetric layer-wise feature guidance}. Finally, inspired by D4RT~\citep{d4rt}, we introduce a target-driven motion decoder that directly estimates each dynamic object's overall relative motion between frames, rather than aggregating point-wise motion (\S~\ref{sec:motion_decoder}). The complete architecture is illustrated in Figure~\ref{fig:pipeline}; we additionally build a separate sky model to handle the sky region independently (see Appendix~\ref{app:sky}).

% \begin{figure}[t]
%     \centering
%     \includegraphics[width=\linewidth]{figure/separator.png}
%     \caption{Architecture of the Motion-Static Separator. Orange, blue, black, and yellow tokens denote static, unprocessed dynamic, actively processed dynamic, and camera parameter tokens respectively. The \textbf{Static Path} (left) applies masked alternating attention to suppress dynamic contamination, followed by a Replace operation. The \textbf{Dynamic Path} (right) applies unmasked alternating attention, then two Deformable Attention branches to produce pure Dynamic Tokens and Mixed Dynamic-Static Tokens.}
%     \label{fig:separator}
% \end{figure}

\subsection{Motion-Static Separator}
\label{sec:separator}
Given an unposed RGB image sequence $\{I^t \mid I^t \in \mathbb{R}^{H \times W \times 3},\, t = 1, \ldots, N\}$, we encode each frame with a frozen DINOv3~\citep{simeoni2025dinov3} encoder to obtain patch token sequences $F^t \in \mathbb{R}^{L \times d}$, where $L$ is the number of patches and $d$ the feature dimension. In parallel, SAM2~\citep{ravi2025sam} produces a binary dynamic mask $M^t \in \{0,1\}^{L}$ for each frame by segmenting from the 2D box of each dynamic object, where $M^t_i = 1$ indicates a dynamic token at position $i$. Based on $M^t$, the token sequence is partitioned into a static subset $F^t_s = \{F^t_i \mid M^t_i = 0\}$ and a dynamic subset $F^t_d = \{F^t_i \mid M^t_i = 1\}$.

The separator processes these tokens through two parallel paths. In the \textbf{Static Path}, the token sequence is processed through three layers of alternating attention~\citep{vggtomega}, following the within-frame and cross-frame design of VGGT-$\Omega$. To ensure the static tokens remain uncontaminated by dynamic content throughout all layers, we impose a masking constraint on each attention computation: the attention logits over dynamic key positions are forced to $-\infty$ before the softmax:
\begin{equation}
  \hat{A}^{(t)}_{i,\,(t',j)} =
  \begin{cases}
    \mathbf{q}^{(t)\,\top}_i \mathbf{k}^{(t')}_j / \sqrt{d} & \text{if } M^{t'}_j = 0, \\
    -\infty & \text{if } M^{t'}_j = 1,
  \end{cases}
  \quad \tilde{F}^{t}_s = \mathrm{softmax}\big(\hat{A}^{(t)}\big)\,V,
  \label{eq:dynamic_mask}
\end{equation}
 where $t$ and $t'$ denote the frames of the query and of the key, respectively. This mask prevents the static path from attending directly to dynamic key positions, thereby reducing dynamic leakage in the refined static tokens. However, the mask is not a hard separator: mixed content may remain in the encoder or be reintroduced through boundary-token fusion, so masking suppresses rather than completely eliminates dynamic information. In the \textbf{Dynamic Path}, all tokens are processed through the same three-layer alternating attention~\citep{vggtomega} without any masking, allowing full cross-frame and within-frame context to flow freely across both static and dynamic positions. At the output stage, two improved deformable attention branches~\citep{zhu2021deformable} are applied. The first branch operates over all dynamic tokens $F^t_d$ to further refine and extract pure dynamic representations. Each dynamic token attends to $P$ neighboring tokens whose positions are predicted by the query via learned offsets:
\begin{equation}
  \hat{\mathbf{f}}_i = W_{\mathrm{proj}}\sum_{k=1}^{P} \alpha_{ik}\,\mathbf{v}_{ik}, \quad
  \alpha_{ik} = \mathrm{softmax}_{k}\!\left(\frac{\mathbf{q}_i^\top \mathbf{k}_{ik}}{\sqrt{d}}\right),
\end{equation}
where $P$ is the number of sampled tokens per query, $\mathbf{q}_i = W_q\mathbf{f}_i$, $\mathbf{k}_{ik}$ and $\mathbf{v}_{ik}$ are key and value features interpolated from the patch token grid at query-predicted offset positions, and $W_{\mathrm{proj}}$ is the output projection. The second branch operates over the boundary subset $\mathcal{B} = \{i \mid M^t_i = 1,\, \exists j \in \mathcal{N}(i) : M^t_j = 0\}$, \emph{i.e.,} tokens within the dynamic mask that also overlap with static regions and therefore carry mixed dynamic-static information. This branch extracts the static component from these mixed tokens, which is then weighted-fused into the corresponding positions in $\tilde{F}^t_s$ from the Static Path, helping recover background information at dynamic-object boundaries rather than discarding these tokens. Separation precedes camera and geometry estimation, so Stage~1 estimates camera parameters and static geometry while direct attention to dynamic key positions is suppressed.
\subsection{Two-stage reconstruction}
\label{sec:recon}
Once the static and dynamic tokens have been separated, we initiate a two-stage reconstruction process: a static reconstruction stage, referred to as Stage~1 below, followed by a dynamic reconstruction stage, referred to as Stage~2 below.

\paragraph{Static Reconstruction}
Stage~1 takes the static tokens $\tilde{F}^t_s$ from the Motion-Static Separator and processes them through $L$ layers of alternating attention~\citep{vggtomega}. Because direct attention to dynamic key positions is masked, the static branch is less exposed to dynamic content and can focus on static scene geometry, although residual dynamic information may remain. At every layer, the camera and static tokens are cached into the \textbf{Tokens Bank}, later reused by Stage~2 to condition dynamic reconstruction. The resulting multi-layer tokens are decoded by three dedicated heads. The \textbf{camera head} refines the camera and register tokens through a cross-frame self-attention trunk to decode each frame's camera-to-world extrinsics and intrinsics. The \textbf{depth head} and \textbf{Gaussian head} instead operate on the patch tokens: both fuse features from several intermediate backbone layers through a shared DPT-style multi-scale fusion pyramid, before a final convolutional layer predicts, at every pixel, a depth value with confidence for the depth head and a set of Gaussian attributes for the Gaussian head, namely opacity, spherical-harmonic color, scale, and rotation. The predicted depth is combined with the camera pose and intrinsics and back-projected to obtain each Gaussian's 3D center in world coordinates, which, with the remaining attributes, instantiates the static 3D Gaussian field.

\paragraph{Dynamic Reconstruction}
Stage~2 reconstructs the dynamic tokens from the Motion-Static Separator. To fully exploit the static features from Stage~1, the clean static tokens and the dynamic tokens are first fused into combined tokens, which serve as the initial input to $L$ layers of alternating attention. After each layer's attention computation, only the dynamic tokens are extracted and fused with the cached static tokens retrieved from the Tokens Bank, to form the input to the next layer; the attention outputs at the camera and static positions are discarded rather than carried forward. We term this \textbf{asymmetric layer-wise feature guidance}: the cached static tokens act as fixed conditioning the dynamic branch attends to but never modifies, so the static representations from Stage~1 are preserved while dynamic tokens draw on their geometric context:
\begin{equation}
  \hat{F}^{(l)}_i =
  \begin{cases}
    F^{(l),\,\text{stage1}}_i & \text{if } M^t_i = 0, \\
    \mathrm{AltAttn}\big(\hat{F}^{(l-1)}\big)_i & \text{if } M^t_i = 1,
  \end{cases}
\end{equation}
where $t$ is the frame to which token $i$ belongs and $F^{(l),\,\text{stage1}}$ is the static token cached at layer $l$ in the \textbf{Tokens Bank}; only the dynamic token positions evolve through the network. We denote the final-layer tokens, fused across all frames, as $F^{(L)}$; besides serving as the global memory bank for the instance-level motion decoder (Section~\ref{sec:motion_decoder}), the dynamic tokens extracted from $F^{(L)}$ at the dynamic-masked positions are decoded by a depth head and a Gaussian head into per-pixel depth and Gaussian attributes. Back-projecting this depth using the camera pose and intrinsics recovered in Stage~1 yields each dynamic Gaussian's 3D center in world coordinates for every frame, which, with the remaining attributes, instantiates the dynamic 3D Gaussian field.

\subsection{Instance-level motion decoder}
\label{sec:motion_decoder}
To model the 4D motion of dynamic Gaussians, we draw inspiration from D4RT~\citep{d4rt} and introduce a motion decoder $D$, a compact cross-attention Transformer that constructs instance queries attending to spatial representations to estimate each instance's motion. After the dynamic Gaussians are reconstructed in Stage~2, for each instance we form a query by adding a learnable embedding of its 2D bounding-box prompt $b_m=(x_1,y_1,x_2,y_2)$ to a discrete timestep embedding of the frame $t_m$ in which it is observed. We then adaptively size a sampling window based on the box dimensions, extract the corresponding RGB patch embedding, and add it to the query, giving the query appearance context beyond the box coordinates.

Each query is decoded through $M$ stacked cross-attention blocks against the global memory bank $F^{(L)}$ (Section~\ref{sec:recon}), formed by fusing the static representations from Stage~1 with the dynamic representations from Stage~2 across every frame, jointly encoding static and dynamic information; with no self-attention among queries, every instance is decoded independently of the others:
\begin{equation}
  \mathbf{q}_m^{(l)} = \mathrm{CrossAttn}\big(\mathbf{q}_m^{(l-1)}, F^{(L)}\big), \quad l = 1, \ldots, M,
\end{equation}
where $\mathbf{q}_m^{(0)}$ is the initial query described above and $\mathbf{q}_m^{(M)}$ is passed to the prediction head below.
The refined query is then passed to an MLP head that predicts the 2D projection of the instance's 3D center, its depth, and its rotation angle. Back-projecting the predicted 2D projection and depth through standard pinhole geometry, using the source frame's camera intrinsics from the Stage-1 output, yields its 3D center in that frame's camera coordinates, which, together with the predicted rotation, gives its 3D pose in the queried frame's local camera space. We then transform this pose into a common world coordinate system using the corresponding frame's predicted camera extrinsics. Querying the same instance at a different frame yields a second pose, independently decoded and likewise transformed into the same world coordinate system, and the rigid motion between the two frames is obtained directly as the difference between these two world-space poses, rather than being regressed as a separate output. This directly predicts object-level pose rather than per-point displacement, making the formulation more robust to occlusion and fast motion.

\subsection{Training Objectives}
\label{sec:loss}
We adopt a three-stage training strategy that mirrors the three-stage reconstruction pipeline (Section~\ref{sec:recon} and Section~\ref{sec:motion_decoder}): Stage~1 supervises the static reconstruction together with camera estimation, Stage~2 supervises the dynamic reconstruction, and Stage~3 supervises the instance-level motion decoder once static and dynamic reconstruction are complete.

\paragraph{Static-region loss.}
In Stage~1, we supervise the static reconstruction with the loss function:
\begin{equation}
  \mathcal{L}_{static} = \lambda_1\, \ell^{static}_{rgb} + \lambda_2\, \ell^{static}_{depth} + \lambda_3\, \ell_{camera} .
  \label{eq:static_loss}
\end{equation}
Here, $\ell^{static}_{rgb}$ is the rendering loss used to supervise the static region, computed between the rendered and ground-truth frames restricted to static pixels; it mainly consists of an $\ell_1$ term, an SSIM term, and an LPIPS perceptual term~\citep{zhang2018unreasonable}. $\ell^{static}_{depth}$ supervises the static point cloud with a log-space $\ell_1$ regression term combined with a multi-scale gradient-consistency term; the supervision is applied in a scale-normalized space, with the ground-truth depth normalized by the scene's median LiDAR depth. $\ell_{camera}$ supervises the camera by comparing the predicted pose encoding to the ground truth with an $\ell_1$ loss on the translation and focal-length components and a quaternion-based cosine loss on the rotation component.

\paragraph{Dynamic-region and motion loss.}
In Stage~2, we train the model to reconstruct the dynamic region. In Stage~3, once static and dynamic reconstruction are complete, we train the instance-level motion decoder. The two objectives are
\begin{equation}
  \mathcal{L}_{dynamic} = \lambda_1\, \ell^{dynamic}_{rgb} + \lambda_2\, \ell^{dynamic}_{depth},
  \qquad
  \mathcal{L}_{motion} = \ell_{motion}.
  \label{eq:dynamic_loss}
\end{equation}
Here, $\ell^{dynamic}_{rgb}$ and $\ell^{dynamic}_{depth}$ take the same form as $\ell^{static}_{rgb}$ and $\ell^{static}_{depth}$ (Eq.~\ref{eq:static_loss}) in Stage~1, but are evaluated only over the dynamic region. $\ell_{motion}$ supervises the motion of each dynamic instance: we compute the $\ell_1$ distance between the rigid motion predicted by the instance-level motion decoder (Section~\ref{sec:motion_decoder}) and the ground-truth relative rotation and translation of that instance between consecutive frames, and average the result over all instances.

%% file: tex/4_Experiments.tex
\vspace{-10pt}
\section{Experiments}
\label{sec:experiments}

\vspace{-6pt}

\subsection{Experimental settings}
\label{sec:exp settings}
 We train the feed-forward model on the original training split of the Waymo Open Dataset~\citep{sun2020scalability}, a widely used autonomous driving benchmark comprising real-world driving logs collected across diverse urban scenarios (798 scenes, approximately 190--200 frames each), and evaluate on the test split (202 scenes). \textbf{To assess generalization}, we conduct zero-shot experiments on both the nuScenes dataset~\citep{caesar2020nuscenes} and the MM-AU traffic accident dataset~\citep{fang2024mmau}. MM-AU is the largest and most fine-grained multimodal accident dataset, comprising 58 accident categories, from which we select one scene per category for evaluation. To measure visual quality, we used standard image quality metrics including PSNR, SSIM~\citep{wang2004image}, and LPIPS~\citep{zhang2018unreasonable}; depth accuracy is measured with depth RMSE (D-RMSE), computed after scale alignment between the predicted and ground-truth depth. Camera pose accuracy is measured with the standard AUC for camera pose estimation~\citep{wang2025vggt} , computed as the area under the curve of the fraction of image pairs whose elative rotation and translation errors fall below angular thresholds of $3^{\circ}$ and $30^{\circ}$ (AUC@3$^{\circ}$ and AUC@30$^{\circ}$). Dynamic motion is evaluated by displacing points on each instance with the predicted motion between adjacent input frames and comparing them with the rigid ground truth derived from box trajectories; we report APD$_{3D}$~\citep{koppula2024tapvid3d}, the average fraction of points whose 3D endpoint error falls below thresholds of $0.1$, $0.3$, $0.5$, and $1.0$\,m. Full implementation and training details are provided in Appendix~\ref{sec:implementation_details}.

\subsection{MAIN RESULTS}
\label{sec:results}
\begin{figure}[t]
  \centering
  \begin{tabular}{c@{\hskip 0.02in} c@{\hskip 0.02in} c@{\hskip 0.02in} c@{\hskip 0.02in} c}
    \scriptsize\textbf{GT} & \scriptsize\textbf{AnySplat} & \scriptsize\textbf{DGGT} & \scriptsize\textbf{StreetForward} & \scriptsize\textbf{Ours}  \\

    \includegraphics[width=0.19\linewidth]{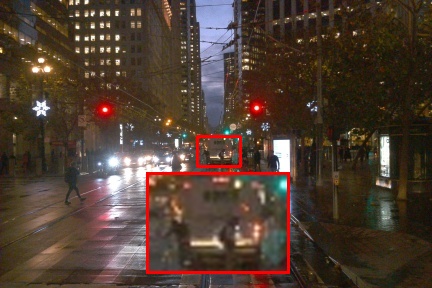} &
    \includegraphics[width=0.19\linewidth]{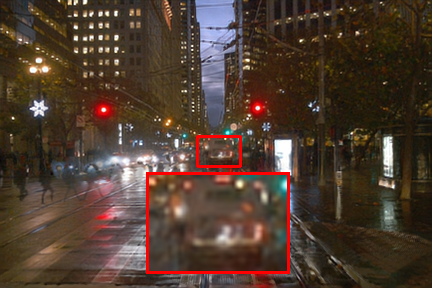} &
    \includegraphics[width=0.19\linewidth]{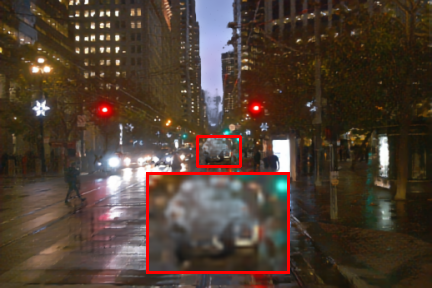} &
    \includegraphics[width=0.19\linewidth]{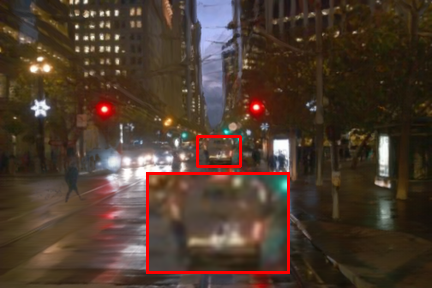} &
    \includegraphics[width=0.19\linewidth]{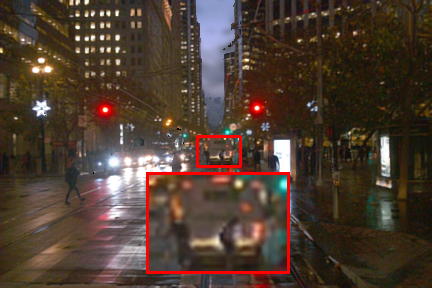} \\ [-0.5mm]

    \includegraphics[width=0.19\linewidth]{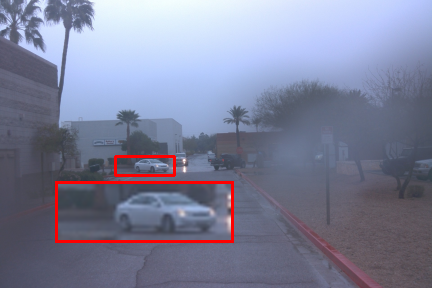} &
    \includegraphics[width=0.19\linewidth]{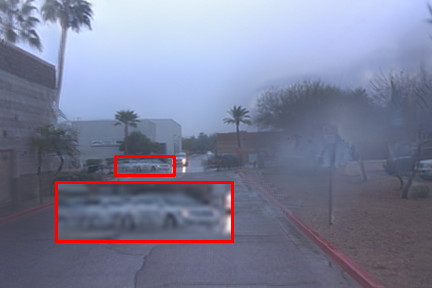} &
    \includegraphics[width=0.19\linewidth]{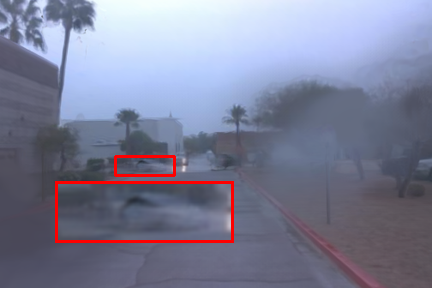} &
    \includegraphics[width=0.19\linewidth]{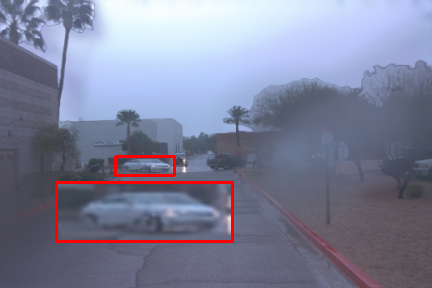} &
    \includegraphics[width=0.19\linewidth]{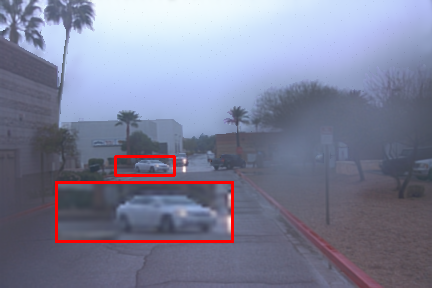} \\ [-0.5mm]

    \includegraphics[width=0.19\linewidth]{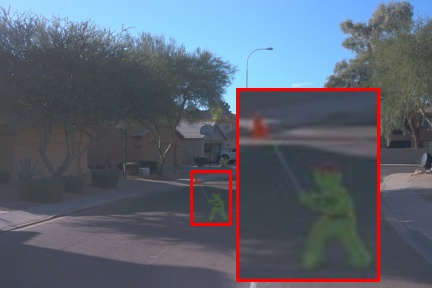} &
    \includegraphics[width=0.19\linewidth]{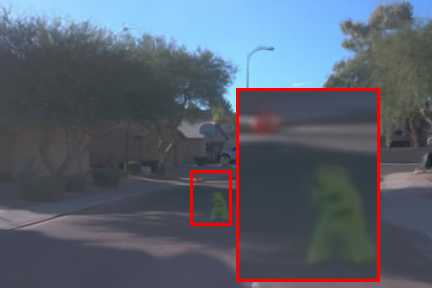} &
    \includegraphics[width=0.19\linewidth]{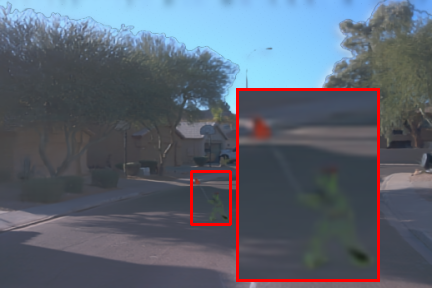} &
    \includegraphics[width=0.19\linewidth]{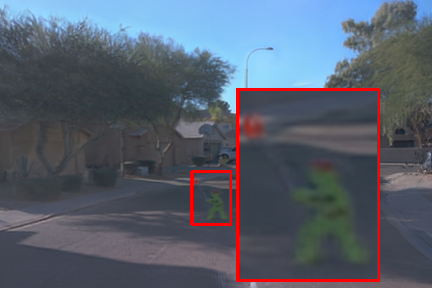} &
    \includegraphics[width=0.19\linewidth]{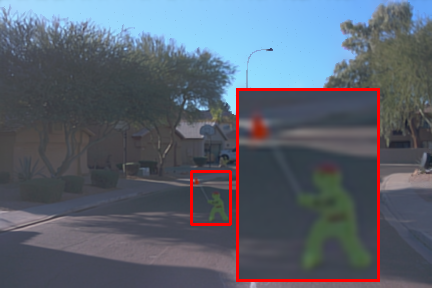} \\ [-0.5mm]
  \end{tabular}
  \caption{\textbf{Qualitative comparison on Waymo.} Compared to AnySplat, DGGT, and StreetForward, our method produces sharper structures and more accurate geometry around dynamic objects, consistent with its lower D-RMSE and higher PSNR/SSIM in Table~\ref{tab:nuscenes_main}.}
  \label{fig:waymo_qual}
\end{figure}

\paragraph{Waymo}
We evaluate in a single-camera setting, using 8 context frames sampled at a stride of 5 and the intermediate frames as target views. We compare against feed-forward baselines for static scenes (MVSplat~\citep{mvsplat}, DepthSplat~\citep{depthsplat}, and AnySplat~\citep{anysplat}) and dynamic scenes (DGGT~\citep{dggt} and StreetForward~\citep{streetforward}); all are evaluated by us under the same setting. Table~\ref{tab:nuscenes_main} reports PSNR, SSIM, and D-RMSE for both full images and dynamic regions. To assess metrics more directly related to static-dynamic decoupling, Table~\ref{tab:waymo_aux} compares S2D-Splat with DGGT and StreetForward on camera-pose AUC, static-region D-RMSE, and APD$_{3D}$. S2D-Splat achieves the best results among the evaluated methods on these metrics while estimating camera poses without pose input. These findings are consistent with the hypothesis that reducing dynamic interference benefits camera estimation, static geometry.
Figures~\ref{fig:waymo_qual} and~\ref{fig:static_dynamic_qual} provide complementary qualitative evidence: compared with the coupled baselines, S2D-Splat yields sharper reconstructions around moving objects and reduces their interference with the static background, preserving background structure in the static render while isolating vehicles in the dynamic render.

\vspace{-6pt}
\begin{table}[t]
  \centering
  \small
  \caption{\textbf{Quantitative Comparison on Waymo.} All methods use 8 input frames, sampled at a stride of 5, as context under our single-camera setting, with intermediate frames as interpolated novel views for evaluation. Higher PSNR/SSIM and lower D-RMSE are better.}
  \label{tab:nuscenes_main}
  \begin{tabular}{lcccccc}
    \toprule
    \multirow{2}{*}{\textbf{Method}} & \multicolumn{3}{c}{\textbf{Full Image}} & \multicolumn{3}{c}{\textbf{Dynamic Region}} \\
    \cmidrule(lr){2-4} \cmidrule(lr){5-7}
    & \textbf{PSNR} $\uparrow$ & \textbf{SSIM} $\uparrow$ & \textbf{D-RMSE} $\downarrow$ & \textbf{PSNR} $\uparrow$ & \textbf{SSIM} $\uparrow$ & \textbf{D-RMSE} $\downarrow$ \\
    \midrule
    MVSplat   & 20.285 & 0.601 & 10.798 & 16.263 & 0.431 & 12.811 \\
    DepthSplat& 21.603 & 0.586 & 9.923 & 17.828 & 0.316 & 11.995 \\
    AnySplat  & 24.146 & 0.761 &7.395 & 19.613 & 0.477 & 9.816 \\
    DGGT      & 24.277 & 0.763 & 4.052 & 19.827 & 0.505 & 6.078 \\
    StreetForward & 25.237 & 0.809 & 3.017 & 20.252 & 0.524 & 5.439 \\
    Ours      & \textbf{25.953} & \textbf{0.814} & \textbf{2.152} & \textbf{22.043} & \textbf{0.719} & \textbf{4.065} \\
    \bottomrule
  \end{tabular}
\end{table}

\begin{table}[t]
  \centering
  \small
  \caption{\textbf{Camera pose, static-region depth, and dynamic motion accuracy on Waymo.} AUC@3$^{\circ}$/AUC@30$^{\circ}$ and APD$_{3D}$ are higher-is-better; Static D-RMSE is lower-is-better.}
  \label{tab:waymo_aux}
  \begin{tabular}{lcccc}
    \toprule
    \textbf{Method} & \textbf{AUC@3$^{\circ}$} $\uparrow$ & \textbf{AUC@30$^{\circ}$} $\uparrow$ & \textbf{Static D-RMSE} $\downarrow$ & \textbf{APD$_{3D}$ (\%)} $\uparrow$ \\
    \midrule
    DGGT      & 0.619 & 0.971 & 3.745 & 13.14 \\
    StreetForward & 0.688 & 0.983 & 2.913 & 11.47\\
    Ours      & \textbf{0.714} & \textbf{0.984} & \textbf{2.057} & \textbf{19.20} \\
    \bottomrule
  \end{tabular}%
  
\end{table}

\paragraph{nuScenes and MM-AU}
We evaluate generalization on nuScenes and MM-AU. For nuScenes, we take the 8 keyframes as input and evaluate on the intermediate non-keyframes; results are shown in Table~\ref{tab:nuscenes_secondary}. For MM-AU, all results in Table~\ref{tab:MM-AU_zero_shot} are obtained zero-shot with the Waymo-trained model: the clips selected for evaluation are all drawn from before the collision, and each clip's 8 frames are fed to the model to evaluate reconstruction under extreme pre-collision driving conditions. Under these conditions, the quality of all baseline methods degrades substantially, whereas our method continues to generalize robustly; we attribute this to our dynamic-static decoupled design, which prevents sudden object motion from degrading reconstruction even in unseen pre-collision scenes. Qualitative comparisons are provided in Appendix~\ref{app:qualitative} (Figures~\ref{fig:motion_qual} and~\ref{fig:mmau_qual}).

\begin{table}[tb]
  \centering
  \small
  \caption{\textbf{Quantitative Comparison under the Zero-Shot Setting on nuScenes.} Our Waymo-trained model shows strong generalization ability in zero-shot NVS on nuScenes.}
  \label{tab:nuscenes_secondary}
  \begin{tabular}{lcccccc}
    \toprule
    \multirow{2}{*}{\textbf{Method}} &
    \multicolumn{3}{c}{\textbf{Scene Reconstruction}} &
    \multicolumn{3}{c}{\textbf{NVS}} \\
    \cmidrule(lr){2-4} \cmidrule(lr){5-7}
     & \textbf{PSNR} $\uparrow$ & \textbf{SSIM} $\uparrow$ & \textbf{LPIPS} $\downarrow$
     & \textbf{PSNR} $\uparrow$ & \textbf{SSIM} $\uparrow$ & \textbf{LPIPS} $\downarrow$ \\
    \midrule
    AnySplat  & 25.136 & 0.837 & 0.165 & 22.204 & 0.701 & 0.289 \\
    DGGT      & 25.352 & 0.862 & \textbf{0.152} & 21.861 & 0.707 &  0.227 \\
    StreetForward & 25.090 & 0.937 & 0.182 & 24.610 & 0.787 & 0.295 \\
    Ours      & \textbf{29.342} & \textbf{0.948} & {0.157} & \textbf{25.503} & \textbf{0.805} & \textbf{0.217} \\
    \bottomrule
  \end{tabular}

\end{table}
\begin{table}[tb]
  \centering
  \small
  \begin{minipage}{0.48\textwidth}
    \centering
    \caption{\textbf{Quantitative Reconstruction Comparison on MM-AU.} Zero-shot results on pre-collision MM-AU clips show our method's strong generalization.}
    \label{tab:MM-AU_zero_shot}
    \begin{tabular}{lccc}
      \toprule
      \textbf{Method} & \textbf{PSNR} $\uparrow$ & \textbf{SSIM} $\uparrow$ & \textbf{LPIPS} $\downarrow$ \\
      \midrule
      AnySplat & 20.150 & 0.754 & 0.316 \\
      DGGT & 20.652 & 0.794 & \textbf{0.189} \\
      StreetForward &  22.150 & 0.920 & 0.328 \\
      Ours & \textbf{25.371} & \textbf{0.932} & {0.204} \\
      \bottomrule
    \end{tabular}
  \end{minipage}
  \hfill
  \begin{minipage}{0.48\textwidth}
    \centering
    \captionof{table}{\textbf{Ablation studies of key components on Waymo.}}
    \label{tab:decoupling_ablation}
    \footnotesize
    \setlength{\tabcolsep}{3pt}
      \begin{tabular}{lccc}
          \toprule
          Method & PSNR $\uparrow$ & SSIM $\uparrow$ & D-RMSE $\downarrow$ \\
          \midrule
          w/o Decoupling & 24.719 & 0.772 & 3.129 \\
          w/o Dyn-Sup & 25.394 & 0.789 & 2.664 \\
          w/o Guidance & 25.521 & 0.795 & 2.539 \\
          Input-only Fusion & 25.637 & 0.806 & 2.297 \\
          2D-Box Mask & 25.660 & 0.809 & 2.387 \\
          w/o Decoder & 25.841 & 0.808 & 2.269 \\
          Ours & \textbf{25.953} & \textbf{0.814} & \textbf{2.152} \\
          \bottomrule
      \end{tabular}
  \end{minipage}
\end{table}
\subsection{Ablation Study}
\label{sec:ablation}

\paragraph{Static-first decoupling.}
To evaluate static-first decoupling as a whole, we remove the entire decoupled design, namely the Motion-Static Separator, the two-stage reconstruction pipeline, and the dedicated depth and Gaussian heads, and replace it with a single coupled baseline: patch tokens from the Feature Extractor are directly processed by one VGGT-$\Omega$ backbone and decoded by a single shared Gaussian head. Since the design is removed as a unit rather than one component at a time, the difference this baseline reports is the aggregate benefit of decoupling; the component-level effects of dynamic suppression, of the guidance mechanism, and of the motion decoder are isolated in the ablations below. Dynamic-object motion then interferes with static scene recovery directly, dropping all metrics substantially (Table~\ref{tab:decoupling_ablation}).

\paragraph{Dynamic suppression.}
Because w/o Decoupling removes several components at once, its performance drop cannot be attributed to any single mechanism. To isolate dynamic suppression, we keep the rest of the architecture unchanged and remove only the $-\infty$ mask on dynamic key positions in Eq.~\ref{eq:dynamic_mask}, allowing static-path attention to attend to dynamic tokens. As shown in Table~\ref{tab:decoupling_ablation}, removing this mask degrades performance: dynamic content can re-enter static features through cross-frame attention, affecting the camera, depth, and Gaussian predictions built on them. This confirms that suppressing dynamic keys contributes to the decoupling gains.

\paragraph{Robustness to mask boundaries.}
To test sensitivity to mask boundaries, we keep the trained model fixed and replace SAM2 masks with 2D detection boxes at test time, labeling a patch as dynamic if any part of it overlaps a box. Although this coarser mask also includes background near box edges, Table~\ref{tab:decoupling_ablation} shows that performance remains close to using SAM2 masks. This robustness likely stems from patch-level tokenization, which already coarsens the partition, and the separator's boundary branch, which recovers static content from mixed tokens.

\paragraph{Asymmetric layer-wise feature guidance.}
Table~\ref{tab:decoupling_ablation} evaluates our asymmetric layer-wise feature guidance, which at every layer fuses the dynamic tokens produced by the current layer with the corresponding cached static tokens retrieved from the Tokens Bank into a complete token set for the next layer. Removing this Tokens Bank fusion causes the dynamic branch to lose access to clean, accurate static priors, so it can no longer exploit this useful information when rendering the dynamic region, visibly lowering the fidelity of that region.

\paragraph{Layer-wise guidance vs. input-only fusion.}
We compare layer-wise guidance with an input-only fusion baseline, where cached static tokens are injected into the dynamic branch only at its input and not in subsequent layers; all other settings remain unchanged. As shown in Table~\ref{tab:decoupling_ablation}, layer-wise guidance performs better than input-only fusion. This gap may arise because layer-wise reinjection allows dynamic tokens at every layer to attend to the corresponding clean static tokens cached from the static branch, which provide a reliable geometric and camera reference during dynamic feature refinement, whereas in the input-only baseline, the static tokens gradually drift away from the accurate static background in subsequent layers.

\paragraph{Instance-level motion decoder.}
To evaluate the motion decoder's contribution, we replace it with a commonly used alternative: instance point clustering combined with TAPIP3D~\citep{zhang2026tapip3d}, an external point-tracking model, to estimate motion instead. Removing the decoder degrades performance across all evaluated metrics (Table~\ref{tab:decoupling_ablation}). This indicates that the motion decoder's direct prediction of instance-level motion is more accurate than the alternative approach of geometric cluster centroids tracked by an external pure-translation tracker, since the latter not only fails to predict rotation but is also prone to being biased by noise points from dynamic objects.

%% file: tex/5_Conclusion.tex
\vspace{-5pt}
\section{Conclusion}
\label{sec:conclusion}

\vspace{-5pt}

We presented \textbf{S2D-Splat}, a pose-free, feed-forward framework for static-first Gaussian reconstruction of dynamic driving scenes. By separating static and dynamic features before camera and geometry estimation, our method reduces dynamic interference during static scene recovery. Cached static features then provide fixed guidance to corresponding layers of the dynamic branch, preserving the established static representations while supplying geometric context for dynamic reconstruction. An instance-level motion decoder further queries the fused scene features to estimate object poses, supporting motion-aligned Gaussian fusion and temporal interpolation. Experiments on Waymo and zero-shot evaluations on nuScenes and MM-AU demonstrate the effectiveness and cross-domain generalization of this design, including on highly dynamic pre-collision clips.

\textbf{Limitations and Future Work.} Our pipeline requires a 2D box prompt for each dynamic instance when standard 3D boxes are unavailable; these prompts must be drawn manually or selected from a 2D detector, which limits scalability. Moreover, reconstruction errors are not fed back to revise the initial instance masks, so segmentation errors may affect both reconstruction stages. Instance-level motion estimation also still has substantial room for improvement. Future work will develop an agent-driven workflow to automate prompt generation or selection, jointly refine masks with reconstruction feedback, and further improve motion estimation.

%% file: tex/6_Appendix.tex
\clearpage
\newpage
\appendix

\section*{\centering\Large\bfseries Appendix}

\section*{Overview}

This appendix provides additional details to complement the main paper:

\begin{itemize}
    \item Details on sky handling and sky modeling are provided in Section~\ref{app:sky}.
    \item The construction of camera poses at unobserved timestamps, together with the motion-aligned Gaussian fusion and deduplication strategy used for interpolated novel-view rendering, is provided in Section~\ref{app:unobserved}.
    \item Implementation details of the two-stage framework, including training configurations and hyperparameters, are provided in Section~\ref{sec:implementation_details}.
    \item Additional qualitative results on nuScenes and MM-AU are provided in Section~\ref{app:qualitative}.
\end{itemize}

\section{Sky Modeling}
\label{app:sky}

The sky has no finite depth and no rigid 3D structure, so folding it into the static or dynamic Gaussian field would force those networks to explain infinite-depth content with a finite-depth representation. We instead exclude the sky entirely from the scene branch, masking the predicted opacity of scene Gaussians to zero at every sky pixel, and model it with a separate sky Gaussian model, following the sky representation of DGGT~\citep{dggt}, shared across all frames in the input window.

Sky Gaussians lie on a sphere of fixed radius $R$ centered at the world origin, with $N{=}90{,}000$ points $\{\mathbf{d}_i\}$ drawn from a Fibonacci sphere sampling scheme for an approximately uniform angular distribution, giving means $\boldsymbol{\mu}_i = R\,\mathbf{d}_i$. Each point's scale is fixed (not learned) from the average distance to its $3$ nearest neighbors on the sphere, so neighboring sky Gaussians overlap smoothly; opacities and orientations are likewise fixed, so only color is estimated from the input images. Each $\boldsymbol{\mu}_i$ is projected into every view of the window, and a sample is valid only if it lands in front of the camera, inside the image bounds, and on a pixel labeled sky by that view's sky mask; the raw color $\bar{\mathbf{c}}_i$ is the average of valid samples across views, and points with no valid sample in any view are discarded. This raw projected color is then refined by a lightweight learned correction MLP $g_\theta$ (two linear layers with a ReLU and a final $\tanh$),
$$
\mathbf{c}_i = \tfrac{1}{2}\tanh\!\big(g_\theta(\bar{\mathbf{c}}_i)\big)_{[:3]} + \tfrac{1}{2} \in [0,1]^3 ,
$$
whose weights are the only trainable parameters of the sky model. Because the sphere is shared across the window, the same sky Gaussians (means, scales, colors) are reused when rendering every frame, differing only in the camera pose used for rasterization.

At render time, the sky Gaussians are rasterized with the camera pose \emph{detached}, so sky gradients cannot affect camera pose estimation, and composited with the scene render $\mathbf{I}_{\text{scene}}$ (accumulated alpha $\alpha_{\text{scene}}$) via a hard gate given by the sky mask $m$:
$$
\mathbf{I} = m\,\mathbf{I}_{\text{sky}} + (1-m)\,\mathbf{I}_{\text{scene}}
$$
This hard, per-pixel gate also decouples optimization: the render loss propagates into the sky correction MLP only through sky-masked pixels and into the static Gaussians only through non-sky pixels, so the two branches never influence the same pixel's loss. The sky model is trained only during Stage~1 of the static reconstruction, supervised solely by its render loss, with sky and scene parameters updated fully independently throughout training.

\section{Camera Poses and Gaussian Fusion at Unobserved Timestamps}
\label{app:unobserved}

\paragraph{Camera pose construction.}
Beyond the world-coordinate transformation described in Section~\ref{sec:motion_decoder}, the camera pose at an unobserved timestamp is constructed from the predicted camera-to-world (c2w) extrinsics of the two adjacent reconstructed keyframes, where the c2w extrinsic denotes the rigid transform that maps points from the camera coordinate frame to the world frame. For a target timestamp $t$ bracketed by keyframe timestamps $t_0$ and $t_1$, we first compute its relative position $\alpha=(t-t_0)/(t_1-t_0)$. The translation is linearly interpolated as
$$
\mathbf{T}_t=(1-\alpha)\,\mathbf{T}_0+\alpha\,\mathbf{T}_1
$$
 The rotation is not interpolated at the matrix level: the endpoint rotation matrices are first projected back onto the space of valid rotations via SVD orthogonalization (with a sign correction on the singular vectors when the determinant is negative), converted to unit-quaternion form, and interpolated with spherical linear interpolation (SLERP) to obtain $\mathbf{R}_t$. The interpolated rotation and translation are then assembled into the target c2w pose, which is inverted to obtain the world-to-camera (w2c) transform required for rasterization. Since the camera poses and the Gaussians are predicted in the same coordinate system, the target camera can directly render the predicted scene without estimating any additional rigid registration.

\paragraph{Motion-aligned Gaussian fusion and deduplication.}
Rendering an unobserved timestamp requires, for each dynamic instance, fusing the Gaussians predicted at the two bracketing keyframes, while removing duplicated copies that would otherwise appear as ghosting. The Gaussians of the two bracketing keyframes are first aligned to the target timestamp using the instance's predicted rigid motion, which consists of a translation offset and a rotation around the instance center, so that the fusion depends only on the instance's relative motion. The motion is applied fractionally in time, forward from keyframe $t_0$ with the fraction $\alpha$ or backward from keyframe $t_1$ with the fraction $1-\alpha$: the Gaussian means are transformed accordingly, and each Gaussian's orientation quaternion is composed with the corresponding rotation via quaternion multiplication, which is equivalent to rotating its covariance as
$$
\boldsymbol{\Sigma}\mapsto R\,\boldsymbol{\Sigma}\,R^{\top},
$$
and keeps anisotropic Gaussians correctly oriented under the warp. Deduplication then follows a nearest-first, visibility-gated scheme: the warped Gaussians from the temporally nearer keyframe ($\alpha<0.5$ selects $t_0$, otherwise $t_1$) constitute the main render, and those from the farther keyframe are composited only at pixels where the accumulated alpha of the main render falls below a hole threshold ($0.2$). The gate distinguishes the two cases that require opposite treatments: surfaces visible in both keyframes are represented solely by the nearer, more reliable copy, while surfaces observed in only one keyframe, such as those occluded in the near view, are still filled by the farther copy, avoiding the missing regions that a hard selection alone would introduce. If the gate is removed and the two sources instead receive complementary time weights $(1-\alpha)$ and $\alpha$, the scheme degenerates into a linear cross-fade of the two reconstructions, in which both copies remain semi-transparent at intermediate times; the visibility-gated selection instead keeps each surface sharp at any $\alpha$.

\section{Implementation Details}
\label{sec:implementation_details}

Each frame is encoded by a frozen DINOv3~\citep{simeoni2025dinov3} image encoder. Training uses the same \textbf{single-camera} input configuration as evaluation: 8 input frames sampled at a stride of 5, with the intermediate frames held out and used as interpolated novel views for supervision. At inference, each dynamic instance only needs a single 2D box in one frame, and SAM2 is prompted with this box to obtain that instance's dynamic mask across all frames; during training, we instead project each instance's annotated 3D bounding box in the training set into a 2D box to speed up data processing. The Motion-Static Separator's alternating-attention backbone consists of 3 layers, while the backbone used in the static and dynamic reconstruction stages consists of 21 layers; both are initialized from frozen VGGT-$\Omega$~\citep{vggtomega} weights and fine-tuned with LoRA adapters. The Gaussian head shares its architecture with the depth head, and the motion decoder consists of 6 cross-attention layers. Training follows a three-stage scheme: Stage~1 trains the static Gaussian reconstruction network while caching intermediate features in a Tokens Bank; Stage~2 trains the dynamic Gaussian reconstruction network conditioned on these static priors once static reconstruction reaches sufficient quality; and Stage~3 trains the motion decoder on the resulting static and dynamic global features to estimate each instance's overall motion. All stages use the AdamW optimizer with a base learning rate of $1\times10^{-4}$ and weight decay of $1\times10^{-4}$, and are trained on $8$ H20 GPUs with a batch size of $8$. The backbone itself remains frozen throughout and is only adapted through its LoRA parameters. All three stages are trained for $50$ epochs each. In Stage~1, the static loss weights (Eq.~\ref{eq:static_loss}) are set to $\lambda_1{=}1.0$, $\lambda_2{=}0.8$, $\lambda_3{=}2.0$, with the rendering term $\ell^{static}_{rgb}$ itself combining an $\ell_1$ term, an SSIM term, and an LPIPS term with weights $1$, $0.5$, and $0.3$, respectively. In Stage~2, the dynamic loss weights (Eq.~\ref{eq:dynamic_loss}) are set to $\lambda_1{=}1.0$, $\lambda_2{=}0.8$, with $\ell^{dynamic}_{rgb}$ using the same $\ell_1$/SSIM/LPIPS weighting as Stage~1.

\paragraph{Box prompts at test time.}
Since all metrics are computed on the test split of each benchmark, the ground-truth 3D bounding boxes can be directly exploited to generate box prompts by projecting them into the image plane. For practical applications where ground-truth 3D annotations are unavailable, we adopt a purely 2D alternative: an off-the-shelf 2D detector such as YOLO first produces all candidate boxes in the video; since objects with small motion amplitude or at long distances have limited influence on reconstruction quality, a human operator can filter out the key moving instances from these candidates; the 2D detection box of a randomly selected frame is then used as the prompt for SAM2~\citep{ravi2025sam}, which generates the instance's 2D masks across all frames.

\paragraph{Target view construction on Waymo.}
In the Waymo evaluation, the target views are the intermediate frames between each pair of adjacent reconstructed keyframes. As part of offline preprocessing, SegFormer~\citep{xie2021segformer} is run on every RGB frame used in the benchmark, including both the reconstruction keyframes and the intermediate target frames; the resulting sky masks are archived by frame index. The keyframe masks are used when constructing the sky model, while the target-view mask is read together with the frame's ground-truth intrinsics and the ground-truth camera pose computed from the ego pose when the intermediate-frame data is loaded, and is used only for sky compositing. The rendering, however, does not directly use the ground-truth pose of the target frame: following Section~\ref{app:unobserved}, the target camera is interpolated from the predicted cameras of the two bracketing reconstructed keyframes according to its timestamp, with the rotation obtained by SLERP and the translation and intrinsics obtained by linear interpolation, which keeps the target camera in the same coordinate system as the predicted Gaussian scene. Consequently, no additional rigid alignment of the target camera is needed: the camera is itself derived by interpolating the predicted camera parameters of the two reconstructed keyframes, and therefore shares the same predicted coordinate frame with the Gaussian field back-projected from those cameras.

In practical deployment, when no pre-generated sky mask is available for the target view, we instead adopt the sky-Gaussian fusion strategy described in Appendix~\ref{app:sky}: sky regions correspond to the locations where the accumulated opacity of the static and dynamic Gaussians is insufficient, and the final image is obtained by blending the scene rendering with the sky rendering at these locations.
\paragraph{Baseline settings.}
Since most of the selected baselines were originally trained on the Waymo dataset, we initialize them directly with their released pretrained weights. For the methods that require adaptation, we fine-tune them following each baseline's own training recipe without any additional architectural modification; their input consists only of multi-frame RGB images, and their motion supervision follows their respective motion modeling.

\section{Additional Qualitative Results}
\label{app:qualitative}
\begin{figure*}[t]
  \centering
  \setlength{\tabcolsep}{2pt}
  \begin{tabular}{c c c c c}
    & \scriptsize\textbf{GT} & \scriptsize\textbf{DGGT} & \scriptsize\textbf{StreetForward} & \scriptsize\textbf{Ours} \\
    \parbox[b][0.14\textwidth][c]{1.3cm}{\centering\scriptsize\textbf{Full}} &
    \includegraphics[width=0.19\textwidth]{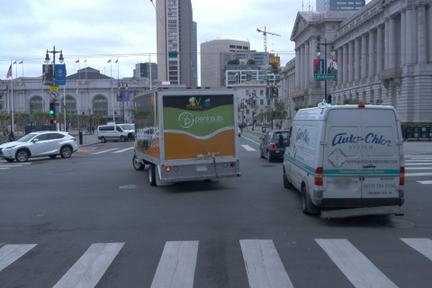} &
    \includegraphics[width=0.19\textwidth]{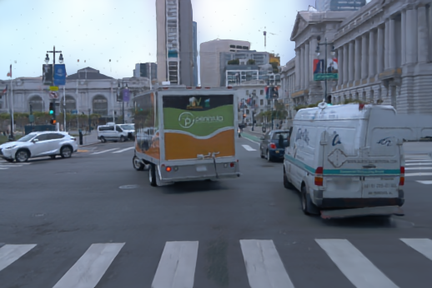} &
    \includegraphics[width=0.19\textwidth]{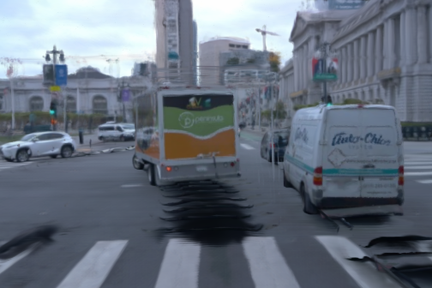} &
    \includegraphics[width=0.19\textwidth]{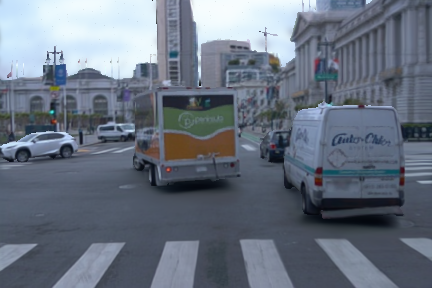} \\
    \parbox[b][0.14\textwidth][c]{1.3cm}{\centering\scriptsize\textbf{Static}} &
    \includegraphics[width=0.19\textwidth]{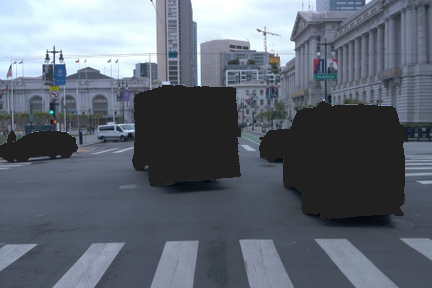} &
    \includegraphics[width=0.19\textwidth]{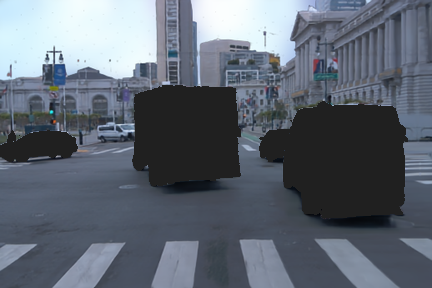} &
    \includegraphics[width=0.19\textwidth]{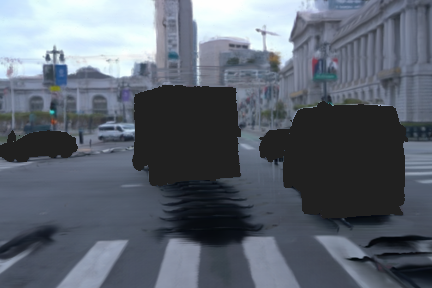} &
    \includegraphics[width=0.19\textwidth]{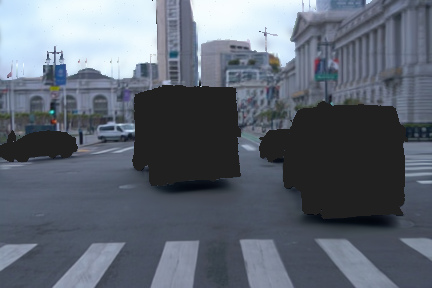} \\
    \parbox[b][0.14\textwidth][c]{1.3cm}{\centering\scriptsize\textbf{Dynamic}} &
    \includegraphics[width=0.19\textwidth]{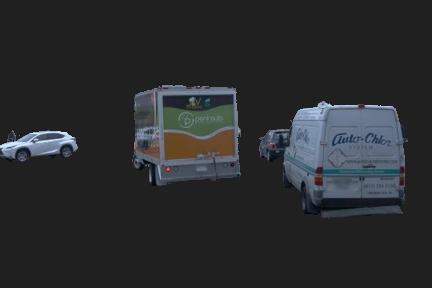} &
    \includegraphics[width=0.19\textwidth]{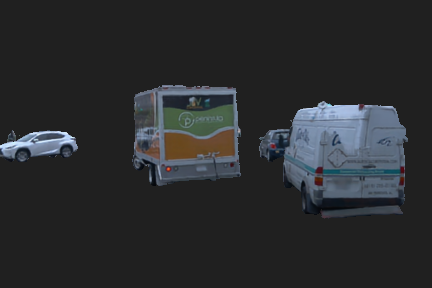} &
    \includegraphics[width=0.19\textwidth]{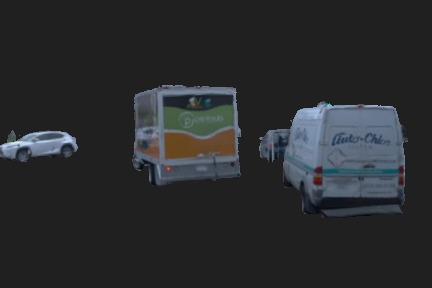} &
    \includegraphics[width=0.19\textwidth]{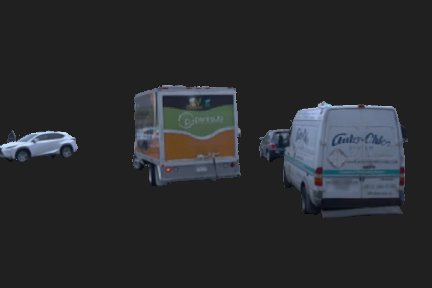} \\
  \end{tabular}
  \caption{Qualitative comparison of scene decomposition on a Waymo example. Columns show ground truth , DGGT, StreetForward, and S2D-Splat; rows show the full reconstruction and the separately rendered static and dynamic components.}
  \label{fig:static_dynamic_qual}
\end{figure*}

\begin{figure}[ht]
  \centering
  \setlength{\tabcolsep}{1pt}
  \begin{tabular}{c c c c}
    \scriptsize\textbf{GT Tracks at $T$} & \scriptsize\textbf{Ours Tracks at $T$} &
    \scriptsize\textbf{GT Flow at $T+1$} & \scriptsize\textbf{Ours Flow at $T+1$} \\
    \includegraphics[width=0.23\linewidth]{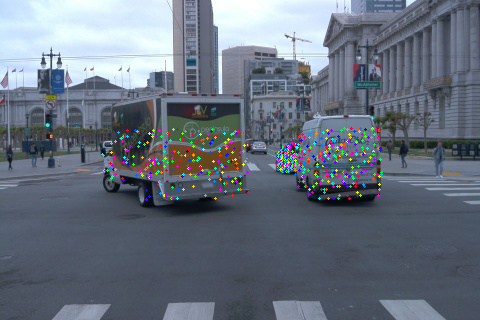} &
    \includegraphics[width=0.23\linewidth]{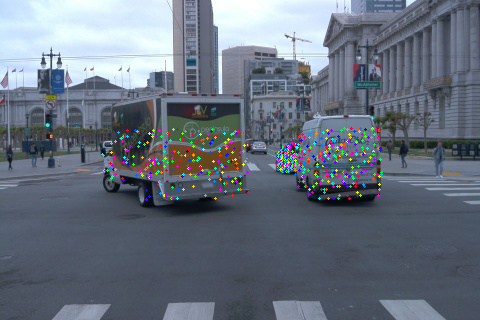} &
    \includegraphics[width=0.23\linewidth]{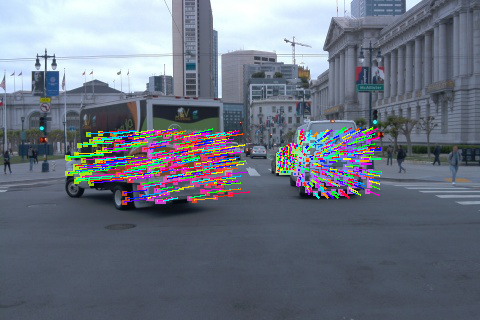} &
    \includegraphics[width=0.23\linewidth]{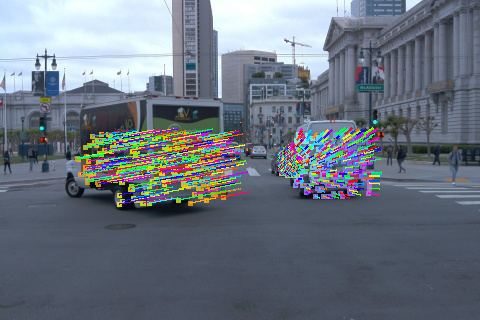} \\
  \end{tabular}
  \caption{Qualitative visualization of motion for two vehicles. The first two panels show ground-truth and predicted tracks at frame $T$; the last two show the corresponding ground-truth and predicted flow at frame $T+1$.}
  \label{fig:motion_flow}
\end{figure}

\begin{figure}[ht]
  \centering
  \setlength{\tabcolsep}{1pt}
  \newcolumntype{L}{>{\centering\arraybackslash}m{1.9cm}}
  \begin{tabular}{L@{\hskip 0.02in}c@{\hskip 0.02in}c@{\hskip 0.02in}c}
    & \scriptsize\textbf{Previous Frame} & \scriptsize\textbf{Interpolated Frame} & \scriptsize\textbf{Next Frame} \\
    \raisebox{-0.5\height}{\mbox{\scriptsize\textbf{Ours}}} &
    \raisebox{-0.5\height}{\includegraphics[width=0.23\linewidth]{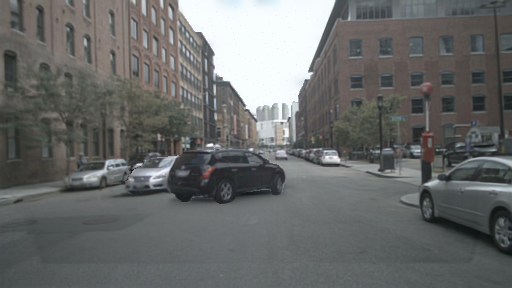}} &
    \raisebox{-0.5\height}{\includegraphics[width=0.23\linewidth]{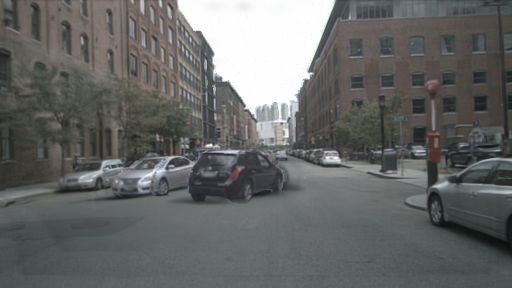}} &
    \raisebox{-0.5\height}{\includegraphics[width=0.23\linewidth]{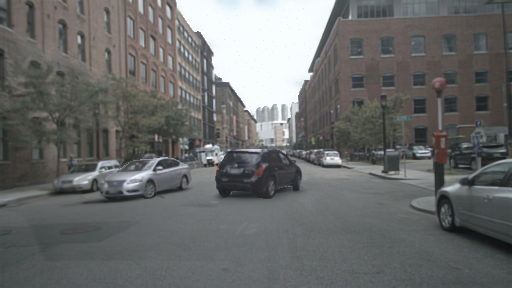}} \\ [-0.5mm]
    \raisebox{-0.5\height}{\mbox{\scriptsize\textbf{StreetForward}}} &
    \raisebox{-0.5\height}{\includegraphics[width=0.23\linewidth]{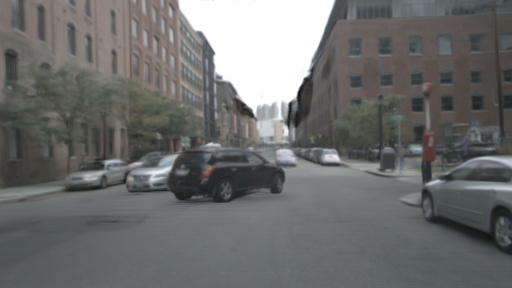}} &
    \raisebox{-0.5\height}{\includegraphics[width=0.23\linewidth]{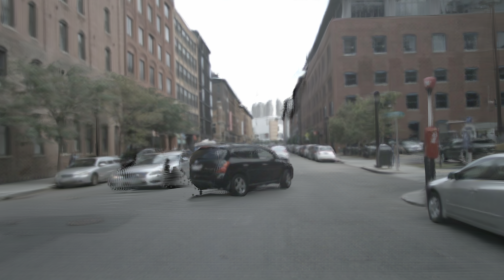}} &
    \raisebox{-0.5\height}{\includegraphics[width=0.23\linewidth]{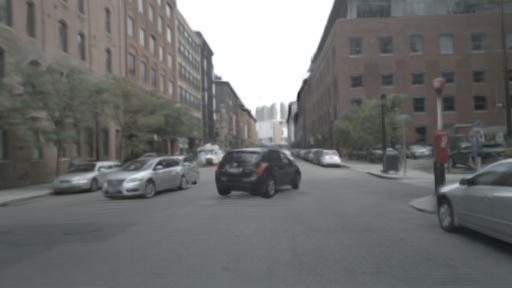}} \\
  \end{tabular}
  \caption{Qualitative comparison of motion interpolation on nuScenes. Left to right: preceding, interpolated, and following frame.}
  \label{fig:motion_qual}
\end{figure}

\begin{figure}[H]
  \centering
  \setlength{\tabcolsep}{1pt}
  \begin{tabular}{c@{\hskip 0.02in}c@{\hskip 0.02in}c}
    \scriptsize\textbf{GT} & \scriptsize\textbf{StreetForward} & \scriptsize\textbf{Ours} \\
    \includegraphics[width=0.28\linewidth]{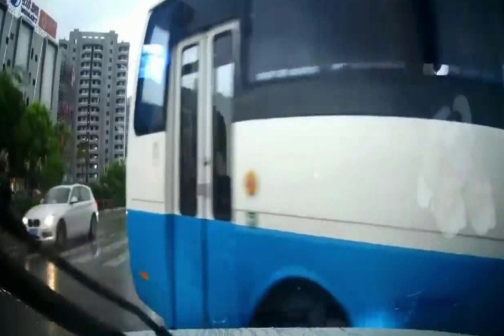} &
    \includegraphics[width=0.28\linewidth]{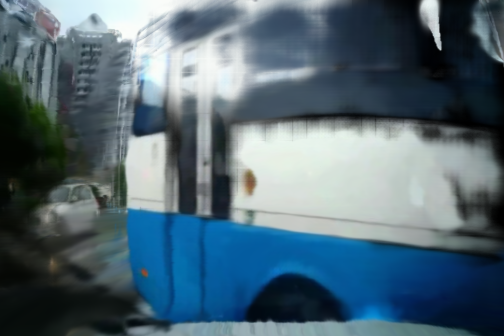} &
    \includegraphics[width=0.28\linewidth]{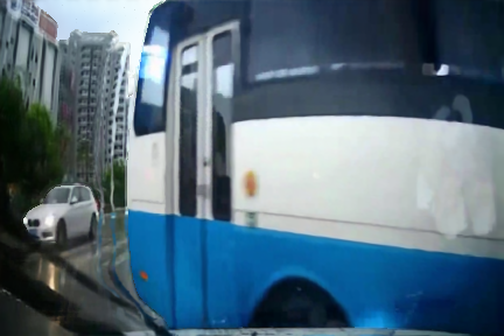} \\ [-0.5mm]
    \includegraphics[width=0.28\linewidth]{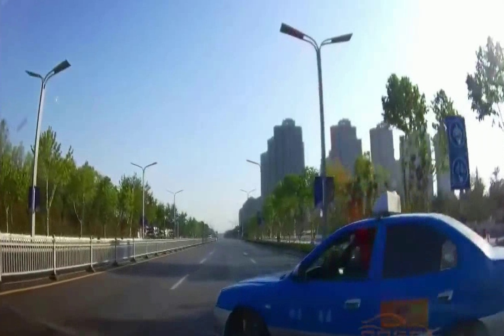} &
    \includegraphics[width=0.28\linewidth]{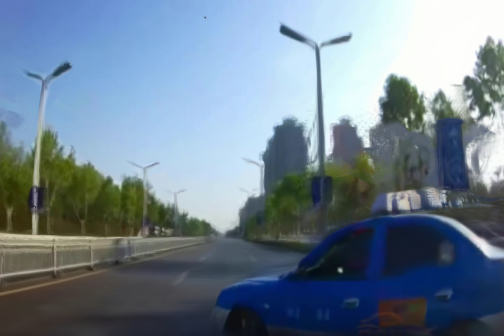} &
    \includegraphics[width=0.28\linewidth]{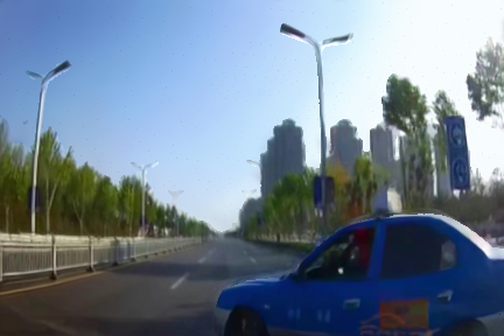} \\
  \end{tabular}
  \caption{Qualitative comparison on MM-AU accident scenes. Left to right: GT, StreetForward, and Ours.}
  \label{fig:mmau_qual}
\end{figure}

Figure~\ref{fig:static_dynamic_qual} first compares full reconstructions with the separately rendered static and dynamic components. The results illustrate that S2D-Splat reduces the interference of highly dynamic objects with static-scene reconstruction: it preserves the background in the static render and isolates vehicles in the dynamic render, with no obvious residual vehicle artifacts, unlike StreetForward and DGGT. Figure~\ref{fig:motion_flow} further visualizes the predicted motion flow and sampled tracks against rigid ground truth for two vehicles. The predicted flow directions and tracks follow the overall reference motion patterns, providing an intuitive view of the decoder's instance-level motion estimates. Figure~\ref{fig:motion_qual} then shows motion interpolation on nuScenes. By integrating an instance's Gaussians across frames into a unified space using the predicted overall motion, our method avoids the large number of independently moving points produced by StreetForward's per-point motion estimation, in which the accumulation of per-point estimation errors leads to inconsistent motion fields; this is particularly evident in the occlusion regions, where StreetForward's interpolated frame has large missing regions.  Finally, Figure~\ref{fig:mmau_qual} shows qualitative results on MM-AU accident scenes.

\clearpage
